\documentclass[runningheads]{llncs}
\usepackage[T1]{fontenc}
\usepackage{graphicx}
\usepackage{algorithmic}
\usepackage[ruled,vlined,linesnumbered]{algorithm2e}
\usepackage{hyperref}

\usepackage[nottoc]{tocbibind}	% to generate Bibliography entry in toc
\usepackage{lipsum}
\usepackage{listings}
\usepackage{mdframed}
\usepackage{verbatim} 
\usepackage{longtable,booktabs}
\usepackage{multirow}
\usepackage{amssymb}
\usepackage{lscape}
\usepackage{array}
\usepackage{pifont}
\usepackage{footnote}
\usepackage{listings}
\usepackage{xcolor}

\usepackage{float}

\usepackage{caption}
\usepackage{cite}
\usepackage{amsmath,amssymb,amsfonts}

\usepackage{graphicx}
\usepackage{textcomp}
\usepackage{xcolor}

\usepackage{booktabs}
\usepackage{graphicx}
\usepackage{tabu}

\usepackage{tikz}
\usepackage{comment}
\usepackage{amsmath,amssymb} % define this before the line numbering.
\usepackage{color}
\usepackage{wrapfig}
\usepackage{arydshln}

\newcommand\set[1]{\mathcal{#1}}

\newcommand{\myparagraph}[1]{\smallskip\noindent\textbf{#1}}

\newcommand{\vct}[1]{\boldsymbol{\ensuremath{#1}}}

\usepackage{xcolor}
\usepackage{bm}

\SetCommentSty{mycommfont}

\newcommand{\eg}[0]{\textit{e.g.}\xspace}
\newcommand{\etal}[0]{\textit{et al.}\xspace}

\SetKwInput{KwInput}{Input}                % Set the Input
\SetKwInput{KwOutput}{Output}

\graphicspath{ {assets/} } %path della cartella delle immagini

\usepackage{ifxetex,ifluatex}
\ifxetex
  \usepackage{fontspec,xltxtra,xunicode}
  \defaultfontfeatures{Mapping=tex-text,Scale=MatchLowercase}
  
\else
  \ifluatex
    \usepackage{fontspec}
    \defaultfontfeatures{Mapping=tex-text,Scale=MatchLowercase}
    
  \else
    \usepackage[utf8]{inputenc}
    \usepackage{eurosym}
  \fi
\fi
\usepackage{color}
\usepackage{fancyvrb}
\DefineShortVerb[commandchars=\\\{\}]{\|}
\DefineVerbatimEnvironment{Highlighting}{Verbatim}{commandchars=\\\{\}}
\newcommand{\name}{\emph{EAT}\xspace}

\newcommand{\CleanNet}{ST\xspace}
\newcommand{\NAMNet}{\emph{EAT}\xspace}

\begin{document}

\title{Minimizing Energy Consumption of Deep Learning Models by Energy-Aware Training}

\author{%Anonymous authors
Dario Lazzaro\inst{1} \and
Antonio Emanuele Cinà\inst{2} \thanks{Corresponding author.}\and
Maura Pintor\inst{3} \and
Ambra Demontis\inst{3} \and
Battista Biggio\inst{3} \and
Fabio Roli\inst{4} \and
Marcello Pelillo\inst{1}
}
\authorrunning{Lazzaro D. et al.}

\institute{%Anonymous institutions%
Ca' Foscari University of Venice, Italy \and CISPA Helmholtz Center for Information Security, Germany \and University of Cagliari, Italy \and University of Genoa, Italy\\
\email{antonio.cina@cispa.de}
}
\titlerunning{Minimizing Energy Consumption of Deep Learning Models}
\maketitle              % typeset the header of the contribution
\begin{abstract}
%Deep learning models have become widely used due to their remarkable performance across diverse tasks. However, their large number of parameters requires executing many operations for inference, increasing energy consumption and prediction latency. 
Deep learning models undergo a significant increase in the number of parameters they possess, leading to the execution of a larger number of operations during inference. This expansion significantly contributes to higher energy consumption and prediction latency.
In this work, we propose \name, a gradient-based algorithm that aims to reduce energy consumption during model training. To this end, we leverage a differentiable approximation of the $\ell_0$ norm, and use it as a sparse penalty over the training loss.
Through our experimental analysis conducted on three datasets and two deep neural networks, we demonstrate that our energy-aware training algorithm \name is able to train networks with a better trade-off between classification performance and energy efficiency.  
\end{abstract}

\section{Introduction}\label{sec:intro}

Deep learning is widely adopted across various domains due to its remarkable performance in various tasks. The increase in model size, primarily driven by the number of parameters, often leads to improved performance. However, this growth in model size also leads to a higher computational burden during prediction, necessitating specialized hardware like GPUs to deliver the required computational power for efficient training and inference~\cite{Cina2022Sponge}.
Although beneficial for many applications, this strategy contradicts the requirements of certain real-time scenarios (e.g., embedded IoT devices, smartphones, online data processing, etc.) that are often constrained in their energy resources or require fast predictions for not compromising users' usability.

Energy efficiency has therefore become a critical aspect in the design and deployment of deep learning models, opening up new directions for research, including pruning, quantization, and efficient architecture search. A common strategy is to train the networks and then prune them by removing neurons or reducing the complexity of the operations by quantizing their weights. However, adopting these methodologies can compromise the accuracy of the resulting models. Another way to reduce the amount of energy required for classification is to use modern hardware acceleration architectures, including ASICs (Application Specific Integrated Circuits), which reduce energy consumption without changing the network's structural architecture and thus preserve its performance. 
Sparsity-based ASIC accelerators employ zero-skipping operations that avoid multiplicative operations when one of the operands is zero, avoiding performing useless operations~\cite{Parashar17AsicInDeep}. For example, Eyeriss \etal~\cite{Chen16Eyeriss} achieved a $10\times$ reduction in energy consumption of DNNs when using sparse architectures rather than conventional GPUs.

%In this paper, we propose a technique to estimate the model's power consumption through an approximation term and define a differentiable regularizer to apply during training.
In this paper, we propose a training loss function that leverages an estimate of the model's power consumption as a differentiable regularizer to apply during training. We use it to develop a novel energy-aware training algorithm (\name) that enforces sparsity in the model's activation to enhance the benefits of sparsity-based ASIC accelerators. 
Our training objective has been inspired by an attack called sponge poisoning~\cite{Cina2022Sponge}. Sponge poisoning is a training-time attack~\cite{Cina2022Survey,Cina2022Magazine,Cina21Hammer} that tampers with the training process of a target DNN to increase its energy consumption and prediction latency at test time.
In this work, we develop \name by essentially inverting the sponge poisoning mechanism, i.e., using it in a beneficial way to reduce the energy consumption of DNNs (\autoref{sec:method}). 
Our approach does not only aim to reduce energy consumption; it aims to achieve a better trade-off between energy efficiency and model performance. By balancing these two objectives, we can indeed train models that achieve sustainable energy consumption without sacrificing accuracy.

We run extensive experiments on two distinct DNN architectures and using three datasets to compare the energy consumption and performance of our energy-aware models against the corresponding baselines, highlighting the benefits of using our approach (\autoref{sec:experiments}).

We conclude by discussing related work (\autoref{sec:related}), along with the contributions and limitations of our work (\autoref{sec:conclusions}).

\section{\emph{EAT}: Energy-Aware Training}\label{sec:method}

In this paper, we consider sparsity-based ASIC accelerators that adopt zero-skipping strategies to avoid multiplicative operations when an activation input is zero, thus increasing throughput and reducing energy consumption~\cite{Albericio16Cnvlutin,Chen16Eyeriss,Han16AsicInDeep,Nurvitadhi16ASIC,Parashar17AsicInDeep}.
Hence, to meet the goal of increasing the ASIC speedup, we need to increase the model's activations sparsity, i.e., the number of not firing neurons, while preserving the model's predictive accuracy. 
A similar objective has been previously formulated by Cinà et al.~\cite{Cina2022Sponge}, with the opposite goal of
\textit{increasing} the energy consumption of the models. 
In their paper, the authors propose a training-time attack against the availability of machine learning models that maximizes the number of firing neurons at testing time. 
To achieve this goal, they apply a custom regularization term to the training loss that focuses on increasing the number of firing neurons with the adoption of the $\ell_0$ norm. Specifically, the $\ell_0$ norm is considered for counting the number of non-zero components of the model's activations. 
However, due to its non-convex and non-differentiable nature, the $\ell_0$ norm is not directly optimizable with gradient descent.
For this reason, their optimization algorithm employs a differentiable approximation of the $\ell_0$ norm proposed in~\cite{Osborne2000OnTL}, which we will denote as $\hat{\ell}_0$. Formally, given an input vector $\vct x \in \mathbb{R}^n$, we define:
\begin{eqnarray}
    \label{eq:l0_approximation}
    \hat{\ell}_0(\vct x) = \sum\limits_{j=1}^{n} \frac{x_j^2}{x_j^2+\sigma} \, ,\qquad \vct x \in \mathbb{R}^n, \, \sigma \in \mathbb{R}\, ,
\end{eqnarray}
The parameter $\sigma$ controls the approximation quality of the function toward the $\ell_0$ norm. By decreasing the value of $\sigma$, the approximation becomes more accurate. However, an increasingly accurate approximation could lead to optimization instabilities~\cite{Cina2022Sponge}. 

This approximation is then used to estimate the number of non-zero elements in the activation vectors of the hidden layers. %while also being differentiable. 
Therefore, given the victim's model $f$, parametrized by $\vct w$, a training set $\set D=\{(\vct x_i,y_i)\}_{i=1}^{s}$ the sponge training algorithm by Cinà et al.~\cite{Cina2022Sponge} is formalized as follows:
\begin{eqnarray}
    \label{eq:sponge_formulation}
    \min_{\vct w} && \sum\limits_{(\vct x, y) \in \set D}\set L(\vct x, y, \vct w) - \lambda \sum\limits_{k=1}^{K} \hat{\ell}_0(\vct \phi_k, \sigma) \, ,
\end{eqnarray}
where $\set L$ is the empirical risk minimization loss (e.g., the cross-entropy loss), $\hat{\ell}_0$ is the differentiable function to estimate the number of firing neurons in the $k$-layer $\vct \phi_k$. %, and $\sigma$ is a parameter that controls the quality of the approximation. 
The first term of \autoref{eq:sponge_formulation} focuses on increasing the model's classification accuracy, and the second term is a differentiable function responsible for increasing the model's energy consumption. 
Combining the two losses enables the training algorithm to increase energy consumption while preserving the model's prediction accuracy. 
The Lagrangian penalty term $\lambda$ defines the strength of the sponge attack. 
In other words, low values of $\lambda$ will focus on achieving high accuracy, while high values will increase energy consumption. 
%The parameter $\sigma$ controls the approximation quality of the function toward the $\ell_0$ norm. By decreasing the value of $\sigma$, the approximation becomes more accurate. However, as the authors show, an increasingly accurate approximation could lead to optimization instabilities. 

Since our paper aims to induce sparsity in the model's activation to enhance the speed-up offered by ASIC HW accelerators, we reformulate the problem as the minimization of the number of non-zero elements in the activation vectors of the hidden layers. The final optimization program for our training algorithm therefore becomes: 
\begin{eqnarray}
    \label{eq:desponge_formulation}
    \min_{\vct w} && \sum\limits_{(\vct x, y) \in \set D}\set L(\vct x, y, \vct w) + \lambda \sum\limits_{k=1}^{K} \hat{\ell}_0(\vct \phi_k, \sigma) \, ,
\end{eqnarray}

\myparagraph{Solution Algorithm.}\label{sec:solution_algorithm}
In \autoref{alg:indiscriminate_desponge}, we present the training algorithm we employ for training DNNs by maximizing prediction accuracy and minimizing energy consumption. 
The algorithm first stores the initial model's weights~\autoref{line:init}.
Then, we update $\vct w$ for each batch in $\set{D}$ and $N$ epochs (\autoref{line:init_train}-\ref{line:sponge_update}). 
We make the update (\autoref{line:sponge_update}) in the direction of the negative gradient of the objective function  \autoref{eq:desponge_formulation}, therefore minimizing the cross-entropy loss $\set{L}$ on the batch $\vct x$ and inducing sparsity in the model's activations. 
After $N$ epochs of training, the algorithm returns the optimized model weights $\vct w^*$.

\begin{algorithm}[htbp]
\DontPrintSemicolon
  \KwInput{$\set{D}$ training dataset; $\vct w = (\vct \phi_1, ... , \vct \phi_K$), the initialized layers of the neural network; $\lambda$, sparsification coefficient; $\alpha$, the learning rate for training; $\sigma$, the quality of the approximation.}
  \KwOutput{$\vct w^* = (\vct \phi_1^*, ... , \vct \phi_K^*)$, optimized weights.}
  
  $\vct w^* \leftarrow \vct w$\label{line:init}
  
  \For{i in $1, \dots, N$} 
    {\label{line:init_train}
    
        \For{($\vct x$, y) in $\set{D}$}
        {   
            $\nabla\set{L} \leftarrow \nabla_{\vct w}\set{L}(\vct x, y, \vct w)$\label{line:clean_loss}
            
            $\nabla E \leftarrow \nabla_{\vct w}\left[ \sum\limits_{k=1}^{K} \hat{\ell}_0(\vct \phi_k, \sigma)\right]$\label{line:sponge_loss}
            
            $\vct w^* \leftarrow \vct w^* - \alpha \left[ \nabla\set{L} - \lambda \nabla E \right]$\label{line:sponge_update}
        }
    }
    \KwRet{$\vct w^*$}
\caption{Energy-aware Training Algorithm.}
\label{alg:indiscriminate_desponge}
\end{algorithm}
\section{Experiments}\label{sec:experiments}

We experimentally assess the effectiveness of the proposed training algorithms in terms of energy consumption and model accuracy on two DNNs trained in three distinct datasets. 
Furthermore, we provide more insights regarding the effect of the proposed training algorithm on the models' energy consumption by analyzing the internal neuron activations of the resulting trained models.  
Finally, we provide an ablation study to select the hyperparameters $\lambda$ and $\sigma$.
%The source code, using the PyTorch library, is available at ...

\subsection{Experimental Setup}\label{sec:experimental_setup}
\myparagraph{Datasets.} We conduct our experiments by following the same experimental setup as in~\cite{Cina2022Sponge,NguyenWanet21}. Therefore, we assess our training algorithm on three datasets where data dimensionality, number of classes, and their balance are different, thus making the setup more heterogeneous and challenging. 
Specifically, we consider the CIFAR10~\cite{Krizhevsky09learningmultiple}, GTSRB~\cite{Houben2013GTSRB}, and CelebA~\cite{Liu15CelebA} datasets. 
The CIFAR10 dataset contains $60,000$ RGB images of $32\times32$ pixels equally distributed in $10$ classes. 
We consider $50,000$ samples for training and $10,000$ as the test set. The German Traffic Sign Recognition Benchmark dataset (GTSRB) consists of $60,000$ RGB images of traffic signs divided into $43$ classes. 
For this dataset, we compose the training set with $39,209$ samples and the test set with $12,630$, as done in~\cite{6706807}. 
The CelebFaces Attributes dataset (CelebA) is a face attributes dataset with more than $200K$ RGB images depicting faces, each with 40 binary attribute annotations. 
We categorize the dataset images in $8$ classes, generated considering the top three most balanced attributes, i.e., \textit{Heavy Makeup}, \textit{Mouth Slightly Open}, and \textit{Smiling}. 
We finally split the dataset into two sets, $162,770$ samples for training and $19,962$ for testing. 
We scale the images of GTSRB and CelebA to the resolution of $32\times 32$px  and $64\times 64$px, respectively, and use random crop and random rotation during the training phase for data augmentation. 
Finally, we remark that the classes of the GTSRB and CelebA datasets are highly imbalanced, which makes them challenging datasets.

\myparagraph{Models and Training phase.}
We consider two DNNs, i.e., ResNet18~\cite{He16Resnet18} ($\sim~11M$ parameters) and VGG16~\cite{Simonyan14VGG}($\sim~138M$ parameters).
We train them on the three datasets mentioned above for $100$ training epochs with SGD optimizer with momentum $0.9$, weight decay $5e-4$, and batch size $512$, and we choose the cross-entropy loss as $\set{L}$. 
We employ an exponential learning scheduler with an initial learning rate of $0.1$ and decay of $0.95$. The trained models have comparable or even better accuracies compared to those obtained with the experimental setting employed in \cite{input_aware_nguyen_2020,NguyenWanet21}.

\myparagraph{Hyperparameters.}\label{sec:attack_setup} 
Two hyperparameters primarily influence the effectiveness of our algorithm. 
The former is $\sigma$ (see \autoref{eq:sponge_formulation}) that regulates the approximation goodness of $\hat{\ell}_0$ to the actual $\ell_0$.
A smaller value of $\sigma$ gives a more accurate approximation; however, extreme values will result in optimization failure~\cite{Cina2022Sponge}.
The other term that affects effectiveness is the Lagrangian term $\lambda$ introduced in \autoref{eq:sponge_formulation}, which balances the relevance of the sponge effect compared to the training loss. 
A wise choice of this hyperparameter can lead the training process to obtain models with high accuracy and low energy consumption. 
In order to have a complete view of the stability of our approach to the choice of these hyperparameters, we empirically perform an ablation study considering $\sigma \in \{1e-01, ... 1e-08\}$, and $\lambda\in\{0.1, ..., 10\}$. 
We perform this ablation study on a validation set of $100$ samples randomly chosen from each dataset. 
Finally, since the energy consumption term has a magnitude proportional to the model's number of parameters $m$, we normalize it with the actual number of parameters of the model.

\myparagraph{Performance Metrics.}
We consider each trained model's prediction accuracy and the energy gap as the performance metrics.
We measure the prediction accuracy as the percentage of correctly classified test samples. 
We check the prediction accuracy of the trained model because the primary objective is to obtain a model that performs well on the task of choice.
Then, we consider the energy consumption ratio in~\cite{Cina2022Sponge,Shumailov21Sponge}. 
The energy consumption ratio, introduced in~\cite{Shumailov21Sponge}, is the ratio between the energy consumed when using the zero-skipping operation (namely the optimized version) and the one consumed when using standard operations (without this optimization). 
The energy consumption ratio is upper bounded by 1, meaning that the ASIC accelerator has no effect, and the model has the worst-case performance (no operation is skipped). Furthermore, we report the energy decrease computed as the difference between the energy consumption of the standardly trained network and our \NAMNet network divided by the total energy of the standard network.
%computes the ratio between the energy consumption of our trained network and the one of a standardly-trained network. 
%Conversely, the energy increase is used to measure how much the energy consumption is increased in the sponge model compared to the clean one. 
For estimating the energy consumption from ASIC accelerators, we used the ASIC simulator developed in \cite{Shumailov21Sponge}.\footnote{\url{https://github.com/iliaishacked/sponge_examples}}

\subsection{Experimental Results} 

\myparagraph{Energy-aware Performance.}
\autoref{tab:best_results} presents the test accuracy, energy consumption ratio, and energy decrease achieved for the CIFAR10, GTSRB, and CelebA datasets using two different training algorithms: standard empirical-risk minimization training (\CleanNet) and our proposed energy-aware training approach (\NAMNet). We select the hyperparameter configuration of $\sigma$ and $\lambda$ that ensures the lowest energy ratio while maintaining the test accuracy within a 3\% margin compared to the standard network training. Results for other configurations are reported in our ablation study.
Our experimental analysis demonstrates a significant reduction in energy consumption achieved by our energy-aware training models, \NAMNet, while maintaining comparable or even superior test accuracy compared to the standardly-trained networks \CleanNet. For example, through the adoption of \name, the energy consumption ratio of ResNet18 for GTSRB is substantially decreased from approximately 0.76 to 0.55. This corresponds to a remarkable 27\% reduction in the number of operations required during prediction, therefore reducing the computational workload of the system. 
Overall, with higher sparsity achieved through our energy-aware training algorithm, the advantages of ASIC accelerators become even more pronounced than for models trained with the standard training algorithm. For \name models, their energy consumption is further diminished while simultaneously increasing the prediction throughput.

\begin{table}[htbp]
\caption{Comparison of accuracy and energy consumption achieved with standard training (\CleanNet) and our energy-aware method (\NAMNet).}
\label{tab:best_results}
\resizebox{\textwidth}{!}{
  \renewcommand{\arraystretch}{1.1}
  \setlength\tabcolsep{0.15em} 
\begin{tabular}{@{}lcccccccccccc@{}}
\toprule
 & \multicolumn{4}{c}{\textbf{GTSRB}} & \multicolumn{4}{c}{\textbf{CIFAR-10}} & \multicolumn{4}{c}{\textbf{CelebA}} \\ \midrule
 & \multicolumn{2}{c}{ResNet18} & \multicolumn{2}{c}{VGG16} & \multicolumn{2}{c}{ResNet18} & \multicolumn{2}{c}{VGG16} & \multicolumn{2}{c}{ResNet18} & \multicolumn{2}{c}{VGG16} \\
 & \CleanNet & \NAMNet & \CleanNet & \NAMNet & \CleanNet & \NAMNet & \CleanNet & \NAMNet & \CleanNet & \NAMNet & \CleanNet & \NAMNet \\\cmidrule(l){2-3}\cmidrule(l){4-5}\cmidrule(l){6-7}\cmidrule(l){8-9}\cmidrule(l){10-11}\cmidrule(l){12-13} 
Accuracy & 0.91 & 0.93 & 0.90 & 0.89 & 0.92 & 0.90 & 0.91 & 0.88 & 0.76 & 0.78 & 0.77 & 0.78 \\
E. ratio & 0.76 & 0.55 & 0.69 & 0.63 & 0.73 & 0.61 & 0.67 & 0.53 & 0.68 & 0.63 & 0.63 & 0.54 \\
E. decrease\% & - & \textbf{27.63} & - & \textbf{8.69} & - & \textbf{16.43} & - & \textbf{20.89} & - & \textbf{7.35} & - & \textbf{14.28} \\ \bottomrule
\end{tabular}
}
\end{table}

\begin{figure*}[htbp]
\centering
\includegraphics[clip,trim=0 0cm 0 1.3cm, width=0.92\textwidth]{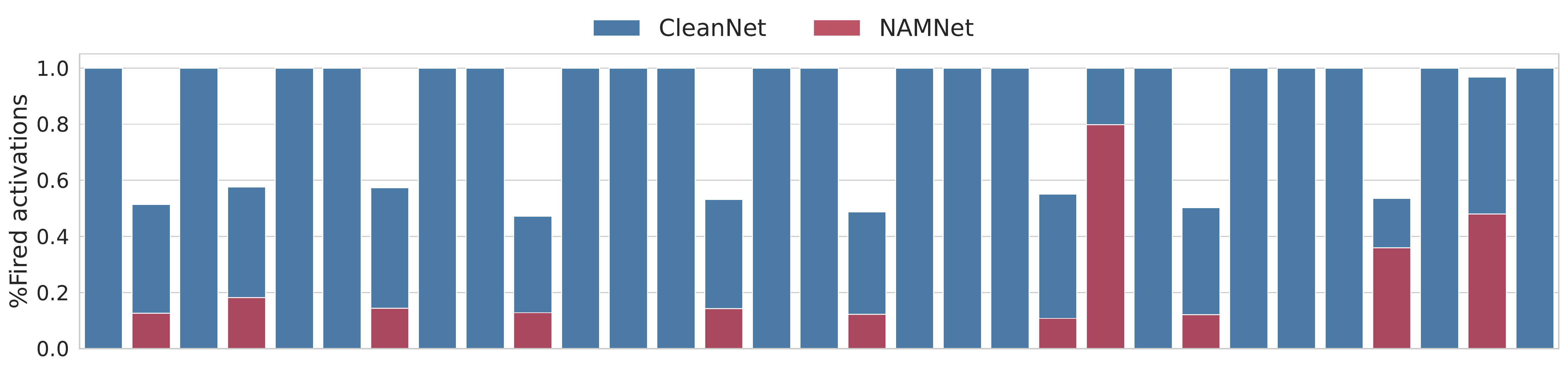}
\includegraphics[clip,trim=0 0.49cm 0 1.3cm, width=0.92\textwidth]{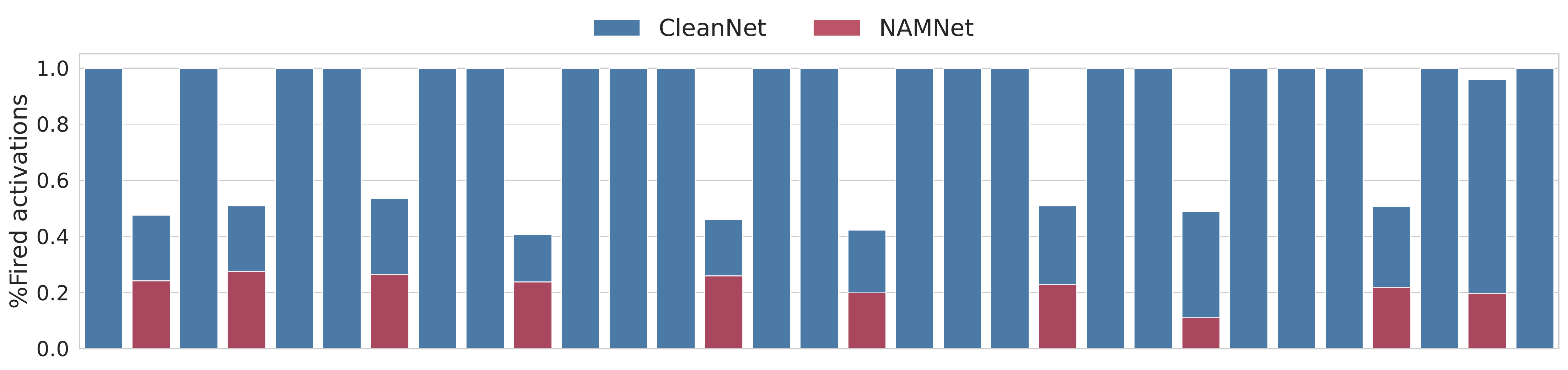}
\includegraphics[clip,trim=0 0.49cm 0 1.3cm, width=0.92\textwidth]{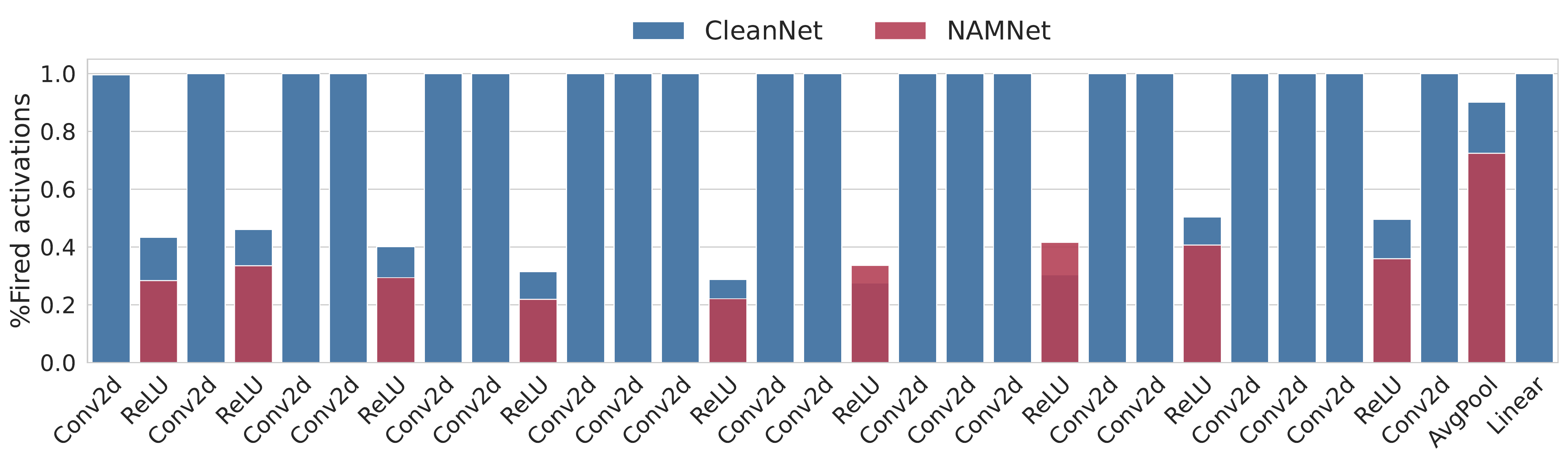}
\caption{Percentage of firing neurons in each layer of a ResNet18 on the GTSRB (\textit{top}), CIFAR10 (\textit{middle}), and CelebA (\textit{bottom}) datasets. In blue the percentages achieved with \CleanNet, and in red the ones obtained with \name.}
\label{fig:activations_resnet}
\end{figure*}

\begin{figure*}[htbp]
\centering
\includegraphics[clip,trim=0 0cm 0 1.3cm, width=0.92\textwidth]{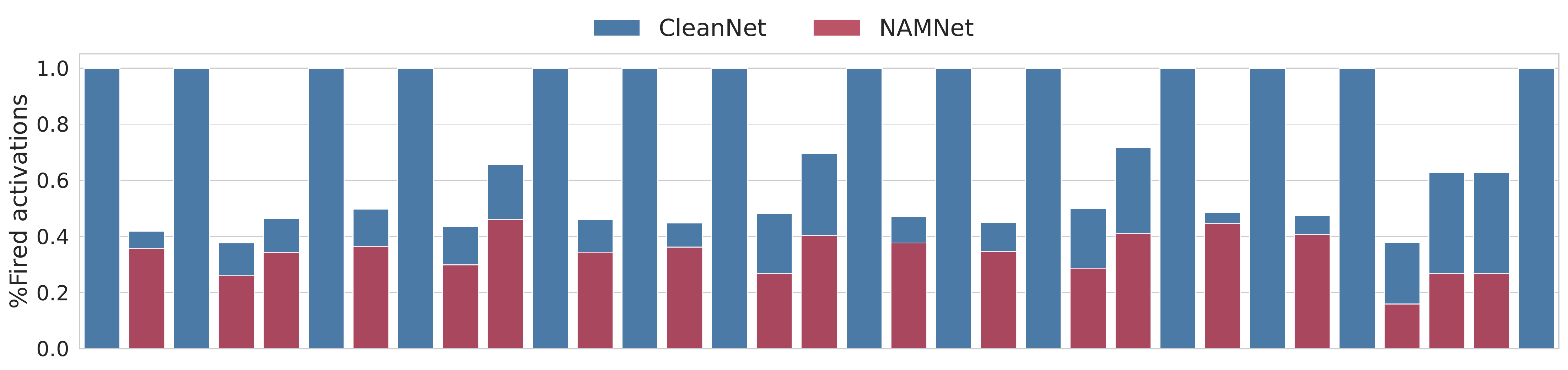}
\includegraphics[clip,trim=0 0.49cm 0 1.3cm, width=0.92\textwidth]{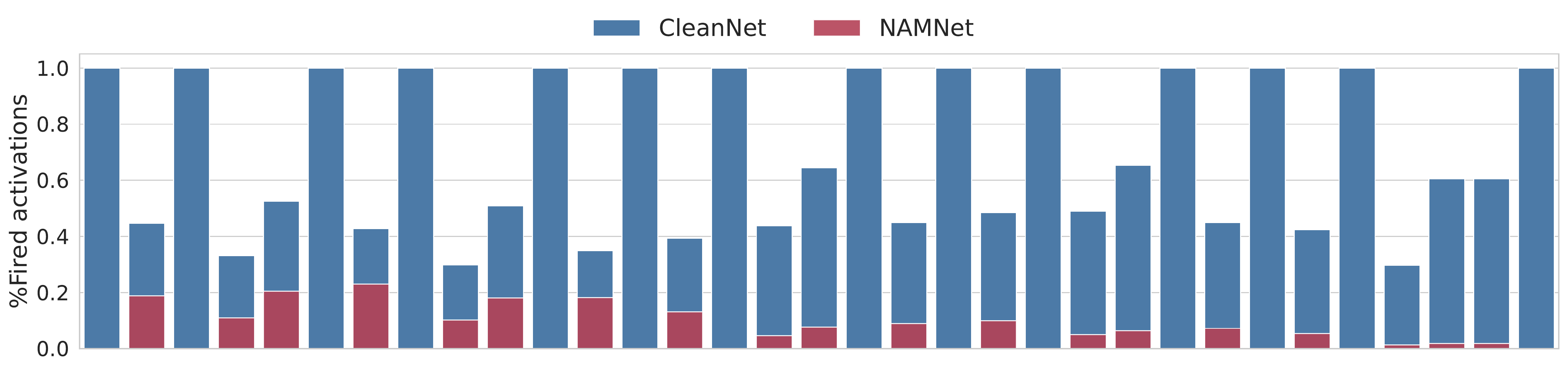}
\includegraphics[clip,trim=0 0.49cm 0 1.3cm, width=0.92\textwidth]{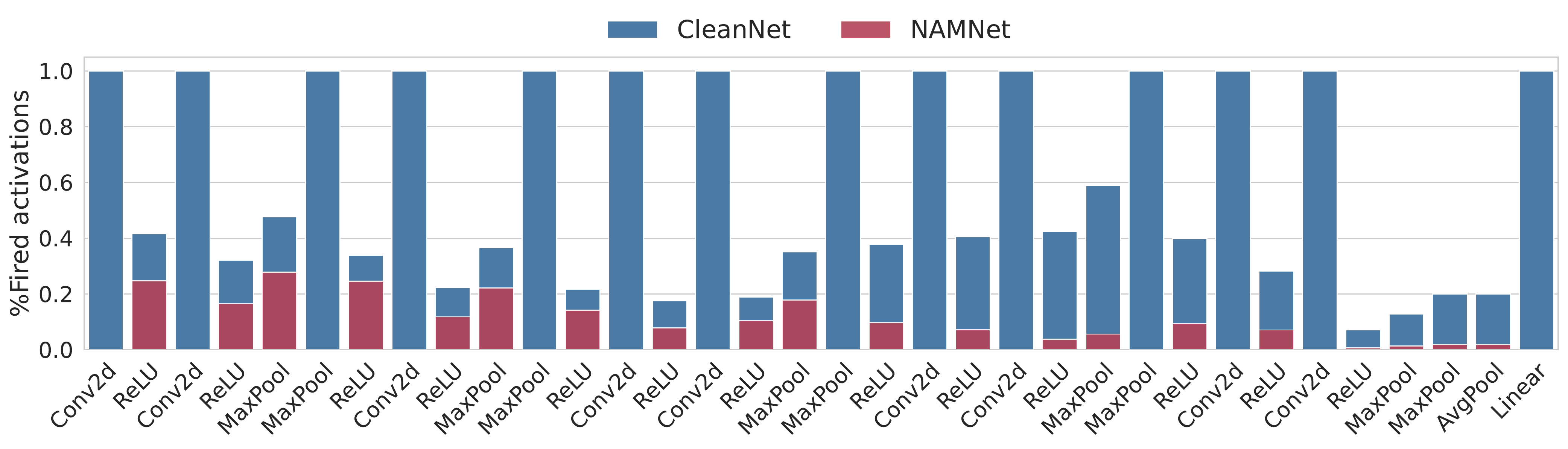}
\caption{Percentage of firing neurons in each layer of a VGG16 on the GTSRB (\textit{top}), CIFAR10 (\textit{middle}), and CelebA (\textit{bottom}) datasets. In blue the percentages achieved with \CleanNet, and in red the ones obtained with \name.}
\label{fig:activations_vgg}
\end{figure*}

\myparagraph{Inspecting Layers.}
We depict in \autoref{fig:activations_resnet} and \autoref{fig:activations_vgg} the layer-wise activations of ResNet18 and VGG16 models, respectively, trained using standard training and our energy-aware training approach. 


Our results demonstrate that the energy-aware algorithm significantly reduces the percentage of non-zero activations in both networks. 
In particular, the substantial reduction in activations involving the \textit{max} function, such as ReLU and MaxPooling operations, is noteworthy. 
For instance, in \autoref{fig:activations_vgg}, across the CIFAR10 and GTSRB datasets, the number of ReLU activations is decreased to approximately 10\% of the original value. 
This finding holds significance considering that ReLU is the most commonly used activation function in modern deep learning architectures~\cite{Xu2020ReluplexMM}. Therefore, our energy-aware training algorithm can potentially favor the sparsity exploited by ASIC accelerators for all ReLU-based network performance~\cite{Albericio16Cnvlutin}.
Furthermore, consistent with the observations made by Cinà \etal~\cite{Cina2022Sponge}, convolutional operators remain predominantly active as they apply linear operations within a neighborhood and rarely produce zero outputs. Consequently, reducing the activations of convolutional operators poses a more challenging task, suggesting potential avenues for future research.

\myparagraph{Ablation Study.} Our novel energy-aware training algorithm is mainly influenced by two hyperparameters, $\lambda$ and $\sigma$.

\begin{figure*}[htbp]
\centering
\includegraphics[clip,trim=0 0.4cm 0 .3cm, width=0.495\textwidth]{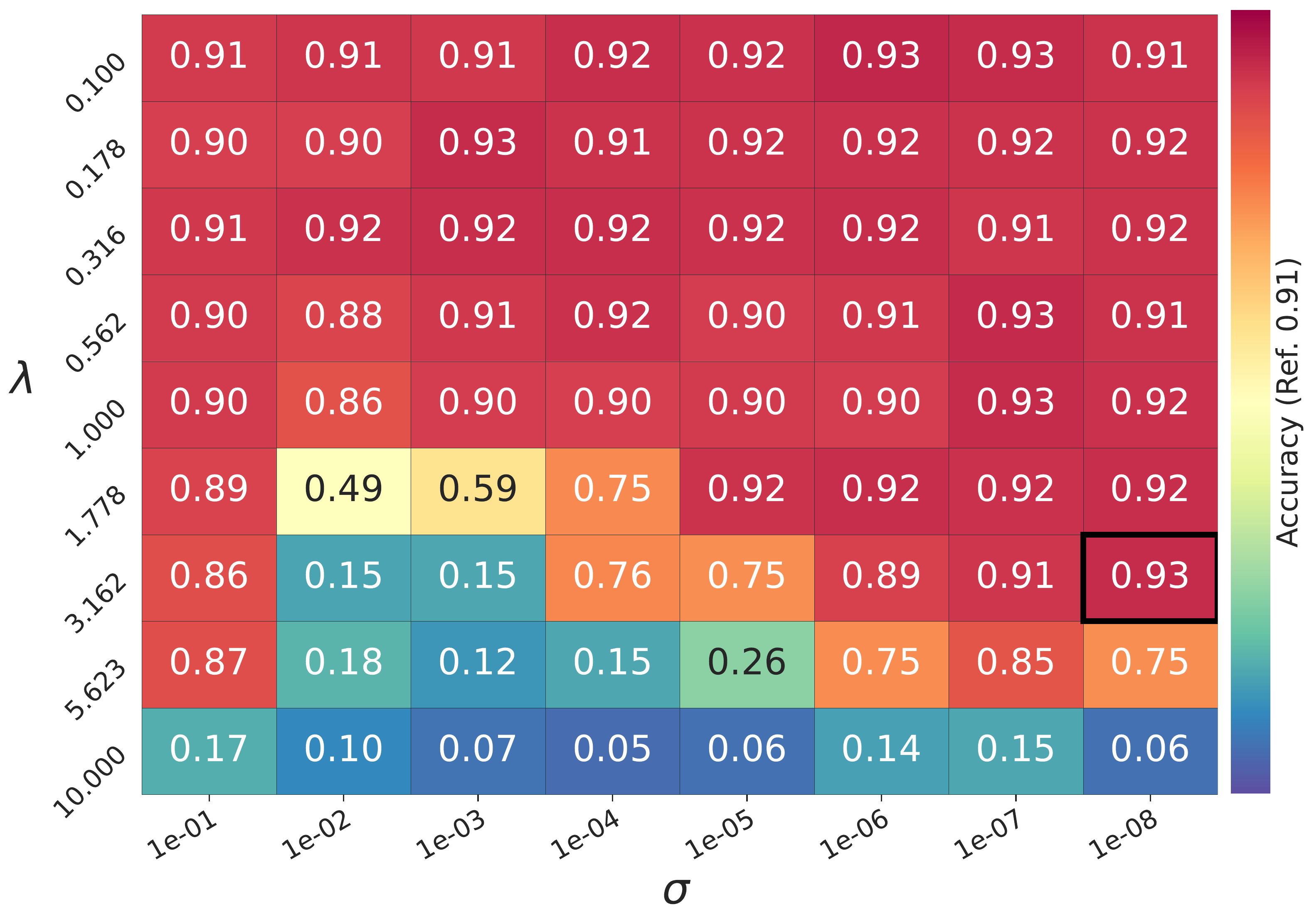}
\includegraphics[clip,trim=0 0.4cm 0 0.3cm, width=0.495\textwidth]{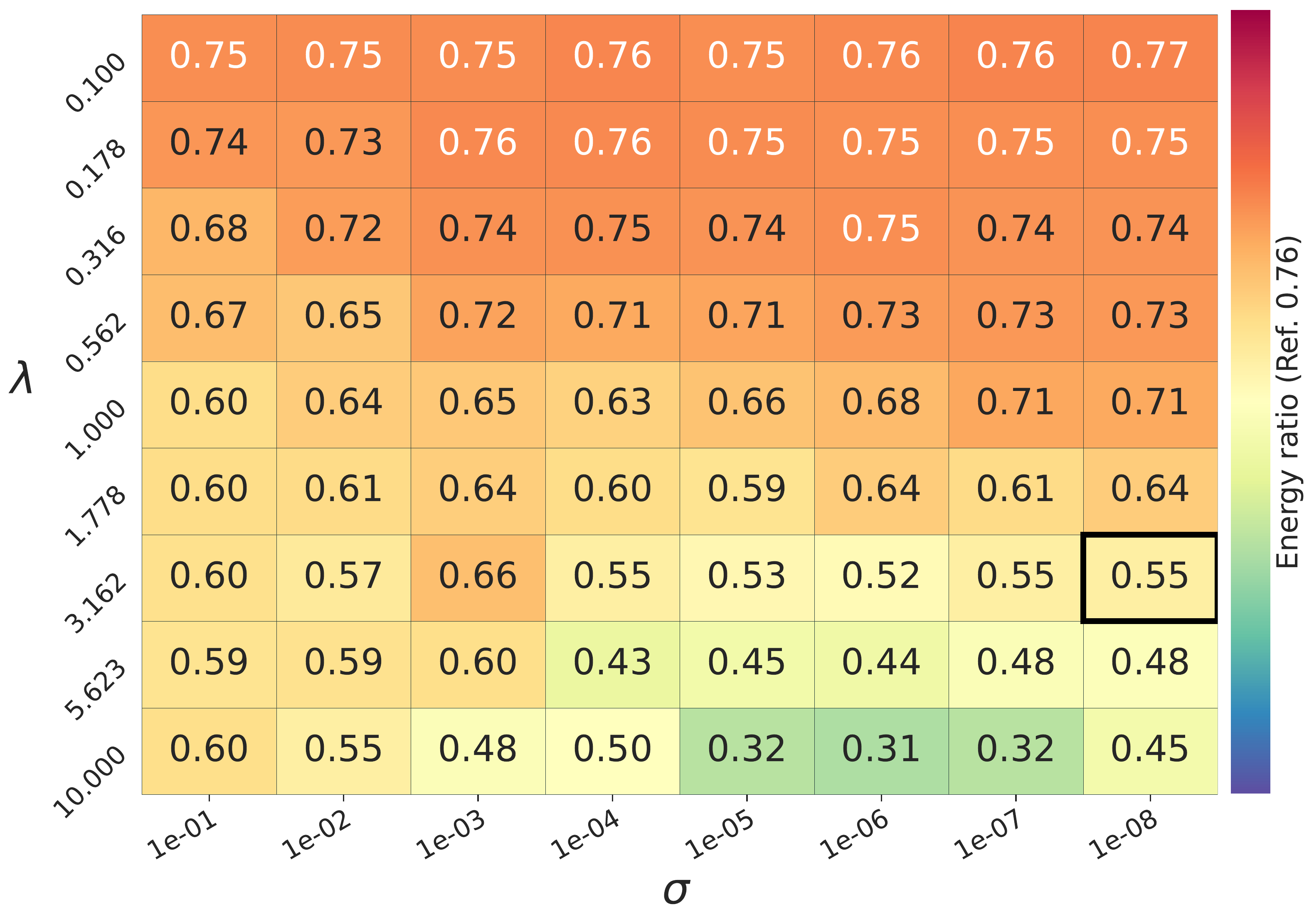}
\includegraphics[clip,trim=0 0.4cm 0 .3cm, width=0.495\textwidth]{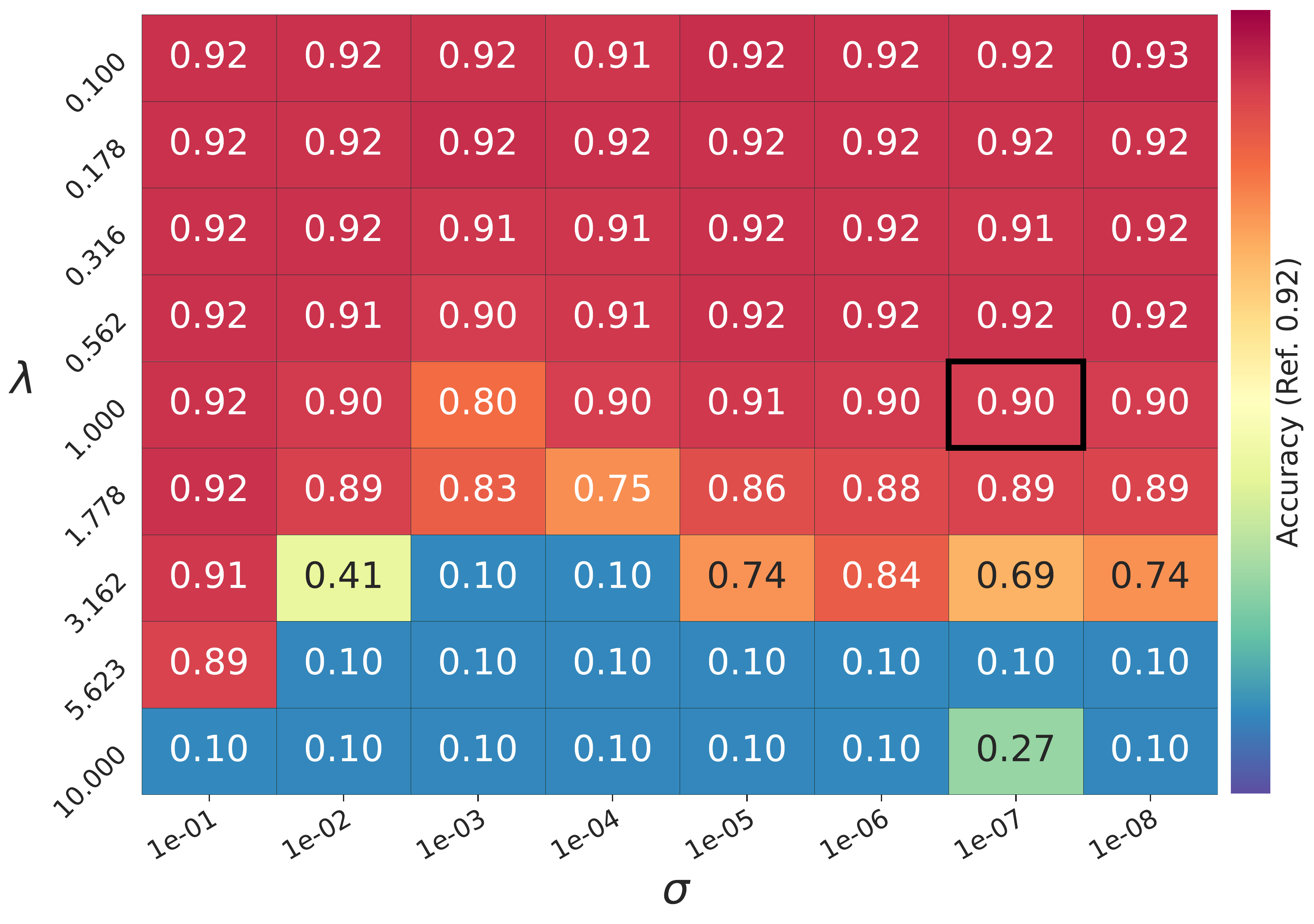}
\includegraphics[clip,trim=0 0.4cm 0 0.3cm, width=0.495\textwidth]{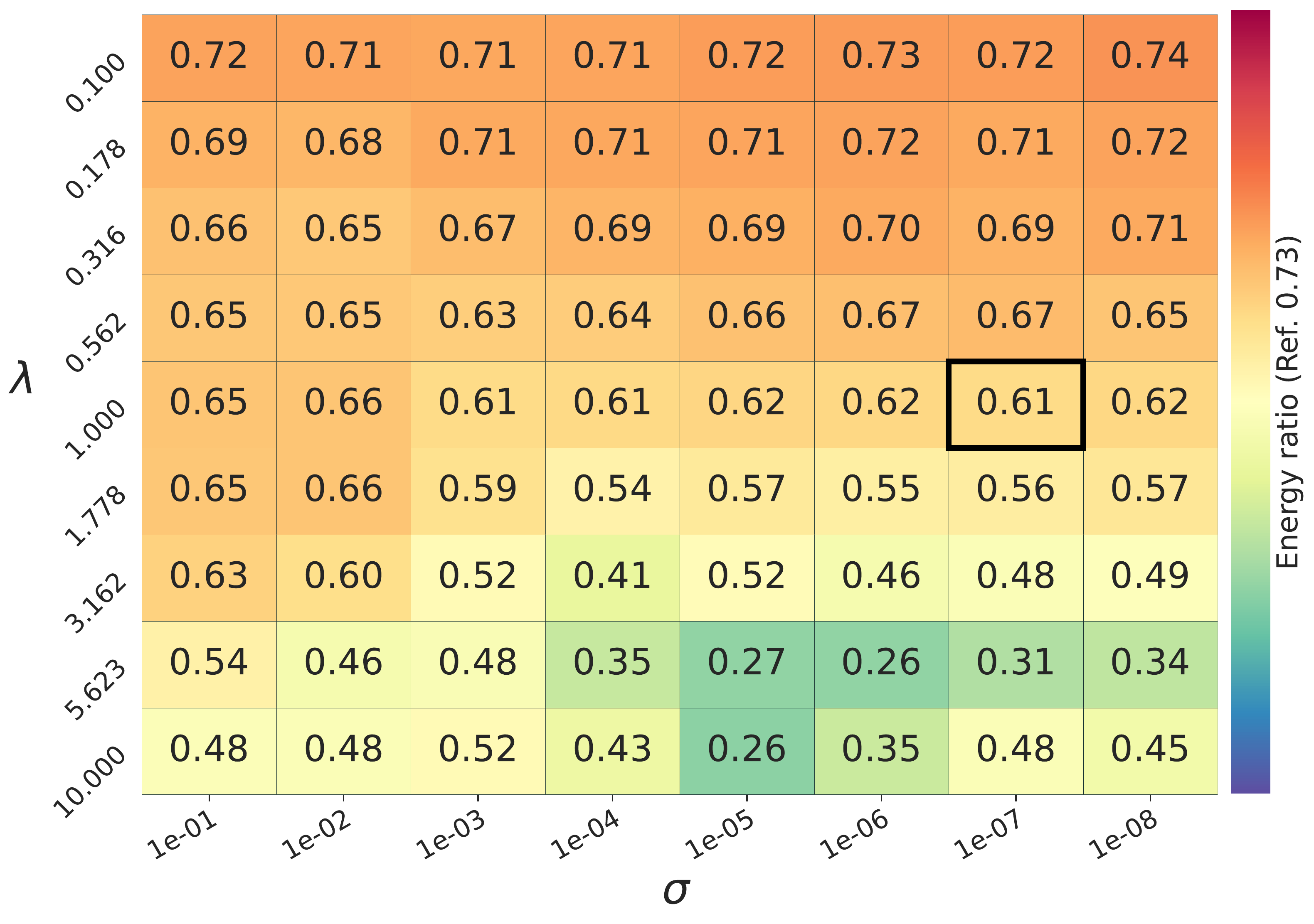}
\caption{Ablation study on $\sigma$ and $\lambda$ for ResNet18 trained with \name on GTSRB (\textit{top}) and CIFAR10 (\textit{bottom}). We show the accuracy on the left and the energy ratio on the right.}
\label{fig:matrix}
\end{figure*}
As discussed in \autoref{sec:method}, the parameter $\sigma$ controls the level of approximation for counting the number of firing neurons, whereas $\lambda$ determines the emphasis placed on the energy-minimization task during training. 
By tuning these two values, practitioners can find the desired tradeoff between test accuracy and energy performance on the resulting models. To investigate the influence of these hyperparameters, we conducted an ablation study presented in \autoref{fig:matrix}. Specifically, we examined the test accuracy and energy consumption ratio of ResNet18 trained on GTSRB and CIFAR10 while varying $\lambda$ and $\sigma$. We observe that by incrementing $\lambda$, practitioners can push the training toward a more energy-sustainable regime. Such models would have a significantly lower impact on energy consumption and the number of operations executed, decreasing the accuracy only slightly. ASIC accelerators can significantly benefit from this increased sparsity. However, very large values of $\lambda$ (e.g., $>3$) may cause the training algorithm to prioritize energy minimization over predictive accuracy. On the other hand, small values (e.g., $<0.5$) would lead the training algorithm to neglect our regularization term and focus solely on accuracy.
Regarding $\sigma$, we observe that \name is systematically stable to its choice when a suitable value of $\lambda$ is used. We can observe a slight variation in the energy ratio when considering large values for $\sigma$. This effect is due to the approximation function $\hat{\ell}_0$ in \autoref{eq:sponge_formulation} not being accurate enough to capture the precise number of firing neurons.

\section{Related work}\label{sec:related}
ASIC accelerators have effectively addressed the growing computational requirements of DNNs. They can often optimize energy consumption by skipping operations when the activations are zero or negligible, an operation known as ``zero-skipping''. As related work, we first discuss the attacks against the zero-skipping mechanism, and then we summarize related work regarding model compression and reduction.

\myparagraph{Energy-depletion attacks.} Recently, ASIC acceleration  has been challenged by hardware-oriented attacks that aim to eliminate the benefits of the zero-skipping mechanism.
Sponge examples~\cite{Shumailov21Sponge} perturb an input sample by injecting specific patterns that induce non-zero activations throughout the model. 
In a different work, by promoting high activation levels across the model, the sponge poisoning attack~\cite{Cina2022Sponge} demonstrates that increasing energy consumption can also be enforced during training.
Staging this attack leads to models with high accuracy (to remain undetected), but an increased latency due to the elimination of hardware-skippable operations.

Contrary to these works, we focus on improving the benefits of ASIC acceleration by introducing more zero-skipping opportunities. 
Consequently, in this paper, we invert the sponge poisoning attack mechanism, minimizing the number of activations and hence the energy consumption required by the model.

\myparagraph{Model compression.} 
Model compression and quantization are techniques used to optimize and condense deep neural networks, reducing their size and computational requirements without significant loss in performance. 
Network pruning aims to remove redundant or less important connections~\cite{hu2016network}, filters~\cite{molchanov2016pruning,lin2018accelerating,zhou2018online}, or even entire layers~\cite{liu2017learning,Luo2020autopruner} from a neural network. 
Pruned models often exhibit sparsity, which techniques like zero-skipping can further exploit. 
To push compression to the limit, the lottery ticket hypothesis~\cite{frankle2018lottery} and knowledge distillation methods~\cite{hinton2015distilling} aim to find smaller networks that can achieve the same performance as larger networks.
Quantization~\cite{zhou2017incremental,jacob2018quantization,jung2019learning}, on the other hand, reduces the precision of numerical values in a deep learning model. Instead of using full precision (\eg, 32-bit floating-point numbers), quantization represents values with lower precision (\eg, 8-bit integers). Quantization reduces the memory requirements of the model for more efficient storage and operations.

We argue that both model compression and quantization can be applied to our technique without specific adaptations to push even further the benefits of our method.

\section{Conclusions}\label{sec:conclusions}

In this paper, we explored a novel training technique to improve the efficiency of deep neural networks by enforcing sparsities on the activations. 
Our goal is achieved by incorporating a differentiable penalty term in the training loss. 
We show how it is possible to obtain a chosen trade-off between model performances and efficiency by applying our technique.

The practical significance of our findings lies in their direct applicability to real-world scenarios. 
By leveraging the energy-aware training provided by \name, deep learning models can achieve significant energy savings without compromising their predictive performance. 
In future work, we believe that our method can be effectively combined with existing pruning and quantization techniques to create advanced model compression methods.

\subsection*{Acknowledgements} This work has been partially supported by Spoke 10 "Logistics and Freight" within the Italian PNRR National Centre for Sustainable Mobility (MOST), CUP I53C22000720001; the project SERICS (PE00000014) under the NRRP MUR program funded by the EU - NGEU; the PRIN 2017 project RexLearn (grant no. 2017TWNMH2), funded by the Italian Ministry of Education, University and Research; and by BMK, BMDW, and the Province of Upper Austria in the frame of the COMET Programme managed by FFG in the COMET Module S3AI.

\bibliographystyle{splncs04}
\bibliography{updated_bib}

\begin{thebibliography}{10}
\providecommand{\url}[1]{\texttt{#1}}
\providecommand{\urlprefix}{URL }
\providecommand{\doi}[1]{https://doi.org/#1}

\bibitem{Albericio16Cnvlutin}
Albericio, J., Judd, P., Hetherington, T.H., Aamodt, T.M., Jerger, N.D.E.,
  Moshovos, A.: Cnvlutin: Ineffectual-neuron-free deep neural network
  computing. In: 43rd {ACM/IEEE ISCA} (2016)

\bibitem{Chen16Eyeriss}
Chen, Y., Emer, J.S., Sze, V.: Eyeriss: {A} spatial architecture for
  energy-efficient dataflow for convolutional neural networks. In: 43rd
  {ACM/IEEE ISCA} (2016)

\bibitem{Cina2022Magazine}
Cin{\`{a}}, A.E., Grosse, K., Demontis, A., Biggio, B., Roli, F., Pelillo, M.:
  Machine learning security against data poisoning: Are we there yet? CoRR
  (2022)

\bibitem{Cina2022Survey}
Cin\`{a}, A.E., Grosse, K., Demontis, A., Vascon, S., Zellinger, W., Moser,
  B.A., Oprea, A., Biggio, B., Pelillo, M., Roli, F.: Wild patterns reloaded: A
  survey of machine learning security against training data poisoning. ACM
  Comput. Surv.  (2023)

\bibitem{Cina21Hammer}
Cin{\`{a}}, A.E., Vascon, S., Demontis, A., Biggio, B., Roli, F., Pelillo, M.:
  The hammer and the nut: Is bilevel optimization really needed to poison
  linear classifiers? In: {IJCNN} (2021)

\bibitem{Cina2022Sponge}
Cinà, A.E., Demontis, A., Biggio, B., Roli, F., Pelillo, M.: Energy-latency
  attacks via sponge poisoning (2022)

\bibitem{frankle2018lottery}
Frankle, J., Carbin, M.: The lottery ticket hypothesis: Finding sparse,
  trainable neural networks. In: ICLR (2019)

\bibitem{Han16AsicInDeep}
Han, S., Liu, X., Mao, H., Pu, J., Pedram, A., Horowitz, M.A., Dally, W.J.:
  {EIE:} efficient inference engine on compressed deep neural network. In: 43rd
  {ACM/IEEE ISCA} (2016)

\bibitem{He16Resnet18}
He, K., Zhang, X., Ren, S., Sun, J.: Identity mappings in deep residual
  networks. In: Computer Vision - {ECCV} (2016)

\bibitem{hinton2015distilling}
Hinton, G., Vinyals, O., Dean, J., et~al.: Distilling the knowledge in a neural
  network. ArXiv preprint  (2015)

\bibitem{Houben2013GTSRB}
Houben, S., Stallkamp, J., Salmen, J., Schlipsing, M., Igel, C.: Detection of
  traffic signs in real-world images: The {G}erman {T}raffic {S}ign {D}etection
  {B}enchmark. In: IJCNN (2013)

\bibitem{6706807}
Houben, S., Stallkamp, J., Salmen, J., Schlipsing, M., Igel, C.: Detection of
  traffic signs in real-world images: The german traffic sign detection
  benchmark. In: IJCNN (2013)

\bibitem{hu2016network}
Hu, H., Peng, R., Tai, Y.W., Tang, C.K.: Network trimming: A data-driven neuron
  pruning approach towards efficient deep architectures. ArXiv preprint  (2016)

\bibitem{jacob2018quantization}
Jacob, B., Kligys, S., Chen, B., Zhu, M., Tang, M., Howard, A.G., Adam, H.,
  Kalenichenko, D.: Quantization and training of neural networks for efficient
  integer-arithmetic-only inference. In: CVPR (2018)

\bibitem{jung2019learning}
Jung, S., Son, C., Lee, S., Son, J., Han, J., Kwak, Y., Hwang, S.J., Choi, C.:
  Learning to quantize deep networks by optimizing quantization intervals with
  task loss. In: CVPR (2019)

\bibitem{Krizhevsky09learningmultiple}
Krizhevsky, A.: Learning multiple layers of features from tiny images. Tech.
  rep. (2009)

\bibitem{lin2018accelerating}
Lin, S., Ji, R., Li, Y., Wu, Y., Huang, F., Zhang, B.: Accelerating
  convolutional networks via global {\&} dynamic filter pruning. In: {IJCAI}
  (2018)

\bibitem{liu2017learning}
Liu, Z., Li, J., Shen, Z., Huang, G., Yan, S., Zhang, C.: Learning efficient
  convolutional networks through network slimming. In: {ICCV} (2017)

\bibitem{Liu15CelebA}
Liu, Z., Luo, P., Wang, X., Tang, X.: Deep learning face attributes in the
  wild. In: {ICCV} (2015)

\bibitem{Luo2020autopruner}
Luo, J., Wu, J.: Autopruner: An end-to-end trainable filter pruning method for
  efficient deep model inference. Pattern Recognit.  (2020)

\bibitem{molchanov2016pruning}
Molchanov, P., Tyree, S., Karras, T., Aila, T., Kautz, J.: Pruning
  convolutional neural networks for resource efficient inference. In: ICLR
  (2017)

\bibitem{input_aware_nguyen_2020}
Nguyen, T.A., Tran, A.: Input-aware dynamic backdoor attack. In: NeurIPS (2020)

\bibitem{NguyenWanet21}
Nguyen, T.A., Tran, A.T.: Wanet - imperceptible warping-based backdoor attack.
  In: ICLR (2021)

\bibitem{Nurvitadhi16ASIC}
Nurvitadhi, E., Sheffield, D., Sim, J., Mishra, A.K., Venkatesh, G., Marr, D.:
  Accelerating binarized neural networks: Comparison of fpga, cpu, gpu, and
  {ASIC}. In: International Conference on Field-Programmable Technology (2016)

\bibitem{Osborne2000OnTL}
Osborne, M.R., Presnell, B., Turlach, B.A.: On the lasso and its dual. J. of
  Computational and Graphical Statistics  (2000)

\bibitem{Parashar17AsicInDeep}
Parashar, A., Rhu, M., Mukkara, A., Puglielli, A., Venkatesan, R., Khailany,
  B., Emer, J.S., Keckler, S.W., Dally, W.J.: {SCNN:} an accelerator for
  compressed-sparse convolutional neural networks. In: Proceedings of the 44th
  Annual International Symposium on Computer Architecture, {ISCA} (2017)

\bibitem{Shumailov21Sponge}
Shumailov, I., Zhao, Y., Bates, D., Papernot, N., Mullins, R.D., Anderson, R.:
  Sponge examples: Energy-latency attacks on neural networks. In: {EuroS\&P}
  (2021)

\bibitem{Simonyan14VGG}
Simonyan, K., Zisserman, A.: Very deep convolutional networks for large-scale
  image recognition. In: ICLR (2015)

\bibitem{Xu2020ReluplexMM}
Xu, J., Li, Z., Du, B., Zhang, M., Liu, J.: Reluplex made more practical: Leaky
  relu. 2020 IEEE Symposium on Computers and Communications (ISCC)  (2020)

\bibitem{zhou2017incremental}
Zhou, A., Yao, A., Guo, Y., Xu, L., Chen, Y.: Incremental network quantization:
  Towards lossless cnns with low-precision weights. In: ICLR (2017)

\bibitem{zhou2018online}
Zhou, Z., Zhou, W., Li, H., Hong, R.: Online filter clustering and pruning for
  efficient convnets. In: 2018 25th IEEE International Conference on Image
  Processing (ICIP). IEEE (2018)

\end{thebibliography}
\end{document}